\documentclass[conference]{IEEEtran}
\usepackage{cite}
\usepackage{amsmath,amssymb}
\usepackage{graphicx}
\usepackage{booktabs}
\usepackage{tikz}
\usepackage{url}
\usetikzlibrary{positioning,arrows.meta}

\begin{document}

\title{Recursive Quantum Long Short-Term Memory for Stable Short-Horizon Temperature Forecasting}

\author{
\IEEEauthorblockN{
Mu-En Lee\IEEEauthorrefmark{1},
Yen-Ku Liu\IEEEauthorrefmark{2},
Samuel Yen-Chi Chen\IEEEauthorrefmark{3},
Yun-Cheng Tsai\IEEEauthorrefmark{4}
}
\IEEEauthorblockA{\IEEEauthorrefmark{1}University of Toronto, Toronto, ON, Canada}
\IEEEauthorblockA{\IEEEauthorrefmark{2}National Yang Ming Chiao Tung University, Hsinchu, Taiwan}
\IEEEauthorblockA{\IEEEauthorrefmark{3}Brookhaven National Laboratory, Upton, NY, USA}
\IEEEauthorblockA{\IEEEauthorrefmark{4}PecuLab LLC, Seattle, WA, USA\\
Yun-Cheng Tsai is the corresponding author: peculab.ai@gmail.com}
}

\maketitle

\begin{abstract}
Quantum long short-term memory (QLSTM) models extend recurrent sequence learning with variational quantum circuits, but their optimization behavior can vary substantially across random initializations and temporal contexts. This paper evaluates a recursive QLSTM architecture against a standard QLSTM for one-step-ahead prediction of daily minimum and maximum temperature. Using daily weather observations from Toronto and identical training settings, we compare convergence, predictive accuracy, and generalization across input windows of 8, 16, and 32 days over 20 random seeds. The recursive model consistently reaches a near-optimal test loss earlier, reduces mean absolute error and root mean squared error, and exhibits a smaller generalization gap. These results indicate that recursive quantum feature transformations can improve stability and out-of-sample performance for compact hybrid quantum--classical temporal models.
\end{abstract}


\section{Introduction}
Long short-term memory (LSTM) networks were designed to model long-range temporal dependencies while mitigating vanishing-gradient behavior in recurrent learning \cite{hochreiter1997long}. Their gated structure has made them a standard baseline for sequential prediction tasks, including meteorological forecasting. Recent quantum machine-learning research has explored hybrid architectures in which trainable variational quantum circuits (VQCs) replace or augment classical transformations \cite{biamonte2017quantum,schuld2019quantum}. Because VQC optimization can be affected by expressibility, initialization, and barren-plateau behavior \cite{mcclean2018barren,abbas2021power}, empirical comparisons should report both accuracy and training stability. One such model, the quantum long short-term memory (QLSTM), embeds quantum neural-network modules in an LSTM-like recurrent structure to learn temporal data using shallow, NISQ-compatible circuits \cite{chen2022qlstm}.

Although QLSTM models can be expressive, their behavior is shaped by circuit initialization, circuit depth, and the amount of temporal context presented to the model. These factors can make both optimization and generalization sensitive to the selected configuration. Recent QLSTM variants have also explored distributed, federated, and fast-weight-style parameter-generation mechanisms for temporal quantum models \cite{chen2025distributedqlstm,chehimi2023fedqlstm,liu2024qtqfwp}. Related quantum sequential-modeling studies have examined shallow quantum temporal embeddings, financial decision systems using LSTM forecasting signals, batch-size/runtime tradeoffs in QLSTM-style training, and sequence-length sensitivity in urban telecommunication forecasting \cite{hsieh2026temporalembeddings,liu2025qrlforecasting,chen2025batchedqlstm,chen2025telecomforecasting}. Recursive QLSTM introduces a metacore-based recursive construction intended to improve temporal information propagation while retaining a compact hybrid architecture \cite{chen2026recursive}. This study provides an empirical comparison of standard QLSTM and Recursive QLSTM for short-horizon temperature prediction.

Our contributions includes: First, evaluating both architectures under identical data, optimization, and quantum-circuit settings; Second, assessing not only MAE and RMSE but also convergence behavior and the generalization gap; Third, reporting mean and standard deviation over 20 random seeds, separating performance trends from single-run randomness.

\section{Background and Experimental Design}
\subsection{QLSTM and Recursive QLSTM}
A conventional LSTM updates its cell state and hidden state through gated transformations of the current input $x_t$ and prior hidden state $h_{t-1}$ \cite{hochreiter1997long}. In QLSTM, selected affine transformations are replaced by VQCs that encode projected features, apply parameterized rotations and entangling operations, and return expectation values to the classical recurrent computation \cite{chen2022qlstm}. This hybrid design can represent nonlinear transformations with a small quantum circuit while leaving training compatible with gradient-based optimization.

Recursive QLSTM augments this design by applying a recursive metacore transformation within the recurrent update. Rather than treating the quantum feature map as a single isolated transformation, the recursive formulation reuses intermediate representations to improve information propagation across the sequence \cite{chen2026recursive}. In this paper, we use the \texttt{MetaCore\_Single\_NN} configuration for the recursive model and hold the remaining architecture settings fixed. See \cite{chen2026recursive} for underlying model and circuit structure.

\subsection{Data and Prediction Task}
We use daily observations from a Toronto weather station for 1 May 2024 through 30 April 2026. Environment and Climate Change Canada provides historical daily weather observations, including temperature variables and degree-day measures, through its public climate-data service \cite{ecccClimate}. The chronological dataset is split into 80\% training observations and 20\% held-out test observations. The task is one-step-ahead prediction of daily minimum and maximum temperature.

Table~\ref{tab:variables} summarizes the model variables. Calendar seasonality is represented by sine and cosine encodings of day of year. All input features are continuous and are scaled using training-set statistics only, preventing test-set information from entering preprocessing.

\begin{table}[t]
\caption{Input and output variables.}
\label{tab:variables}
\centering
\footnotesize
\begin{tabular}{p{0.10\columnwidth}p{0.57\columnwidth}p{0.18\columnwidth}}
\toprule
\textbf{Role} & \textbf{Variable} & \textbf{Unit/scale}\\
\midrule
Input & Day-of-year cosine; day-of-year sine & $[-1,1]$\\
Input & Year number & Count\\
Input & Mean, minimum, and maximum temperature & $^\circ$C\\
Input & Heating and cooling degree days & $^\circ$C\\
Output & Next-day minimum and maximum temperature & $^\circ$C\\
\bottomrule
\end{tabular}
\end{table}

\subsection{Protocol and Metrics}
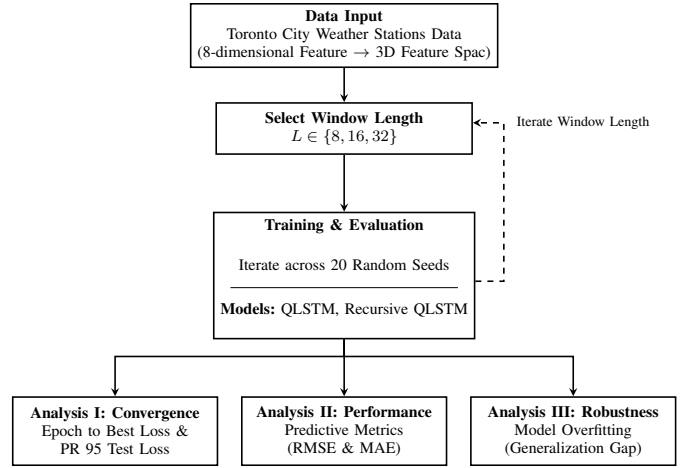
\begin{figure}[htbp]
\centering
\resizebox{0.5\textwidth}{!}{ 
    \begin{tikzpicture}[scale=0.5, node distance=0.6cm and 0.5cm, every node/.style={fill=white, font=\footnotesize}]
    
\tikzset{
    box/.style={rectangle, draw=black, thick, minimum width=4.5cm, minimum height=0.9cm, text centered, align=center},
    resbox/.style={rectangle, draw=black, thick, minimum width=3.6cm, minimum height=1.2cm, text centered, align=center},
    arrow/.style={thick, ->, >=stealth}
}


\node (data) [box] {\textbf{Data Input} \\ Toronto City Weather Stations Data \\ (8-dimensional Feature $\rightarrow$ 3D Feature Spac)};

\node (win) [box, below=of data] {\textbf{Select Window Length} \\ $L \in \{8, 16, 32\}$};

\node (training) [box, below=1.0cm of win, minimum height=2.2cm] {
    \textbf{Training \& Evaluation} \\
    \vspace{2pt} \\
    Iterate across 20 Random Seeds \\
    \vspace{2pt} \rule{4.0cm}{0.4pt} \vspace{2pt} \\
    \textbf{Models:} QLSTM, Recursive QLSTM \\
};

\node (res2) [resbox, below=1.0cm of training] {\textbf{Analysis II: Performance} \\ Predictive Metrics \\ (RMSE \& MAE)};
\node (res1) [resbox, left=0.4cm of res2] {\textbf{Analysis I: Convergence} \\ Epoch to Best Loss \& \\ PR 95 Test Loss};
\node (res3) [resbox, right=0.4cm of res2] {\textbf{Analysis III: Robustness} \\ Model Overfitting \\ (Generalization Gap)};

\draw [arrow] (data) -- (win);
\draw [arrow] (win) -- (training);

\draw [arrow] (training.south) -- ++(0,-0.6) -| (res1.north);
\draw [arrow] (training.south) -- (res2.north);
\draw [arrow] (training.south) -- ++(0,-0.6) -| (res3.north);

\draw [arrow, dashed] ([yshift=-0.2cm]training.east) -- ++(1.0,0) |- ([yshift=0.2cm]win.east) node[midway, right, align=center, font=\scriptsize, xshift=0.1cm] {Iterate Window Length};

\end{tikzpicture} 
}
\caption{Experimental workflow for comparative study across QLSTM and Recursive QLSTM.}
\label{fig:experiment_flowchart}
\end{figure}

Figure~\ref{fig:experiment_flowchart} illustrates the overall experimental workflow.
The models receive a temporal window $L\in\{8,16,32\}$ and predict the following day. Each configuration is trained for 60 epochs using a batch size of 8 and learning rate $5\times10^{-4}$. To characterize stochastic training behavior, each setting is repeated for 20 seeds. Core settings are listed in Table~\ref{tab:params}.

\begin{table}[t]
\caption{Shared experimental configuration.}
\label{tab:params}
\centering
\footnotesize
\begin{tabular}{p{0.53\columnwidth}p{0.35\columnwidth}}
\toprule
\textbf{Parameter} & \textbf{Value}\\
\midrule
QNN depth; hidden size; input projection size; number of qubit & 1; 3; 3; 6\\
Window length $L$ & $\{8,16,32\}$\\
Prediction horizon; input/output dimensions & 1; 8/2\\
Epochs; batch size; learning rate & 60; 8; 0.0005\\
Random seeds & 0--19 (20 trials)\\
Recursive metacore & \texttt{MetaCore\_Single\_NN}\\
\bottomrule
\end{tabular}
\end{table}

We report mean absolute error (MAE), root mean squared error (RMSE), and the generalization gap, defined as test loss minus training loss. A smaller generalization gap indicates a smaller train--test discrepancy. We additionally report the epoch of the minimum test loss (Best Epoch) and the first epoch that reaches 95\% of each run's total improvement ($t_{95}$). The latter gives a practical indicator of how quickly a run obtains a near-optimal test result.

For a test set of $n$ daily forecasts with two temperature targets, let $y_{ij}$ and $\hat{y}_{ij}$ denote the observed and predicted values for forecast $i$ and target $j$, where $j \in \{1,2\}$ corresponds to next-day minimum and maximum temperature. Aggregate MAE and RMSE are computed over both targets and all held-out test observations as

\begin{equation}
\mathrm{MAE}
=
\frac{1}{2n}
\sum_{i=1}^{n}
\sum_{j=1}^{2}
\left|y_{ij}-\hat{y}_{ij}\right|.
\end{equation}

\begin{equation}
\mathrm{RMSE}
=
\sqrt{
\frac{1}{2n}
\sum_{i=1}^{n}
\sum_{j=1}^{2}
\left(y_{ij}-\hat{y}_{ij}\right)^2
}.
\end{equation}

For target-wise analysis, the same metrics are computed separately for minimum and maximum temperature. We also report the 95th-percentile absolute error (P95 AE), coefficient of determination ($R^2$), and absolute bias. Absolute bias is defined as the absolute value of the mean signed prediction error across the held-out test set. For each seed, predictive metrics are calculated at the epoch with the minimum test loss for that run.
For each seed, $t_{95}$ is measured relative to that run's initial and best test losses. This definition is preferable to a fixed loss threshold because the two architectures and their random initializations can begin at different loss levels. All reported error bars represent the standard deviation across the 20 independent runs.

\section{Results}
\subsection{Convergence Behavior}

\begin{figure}[htbp]
\centering
\includegraphics[scale=0.35]{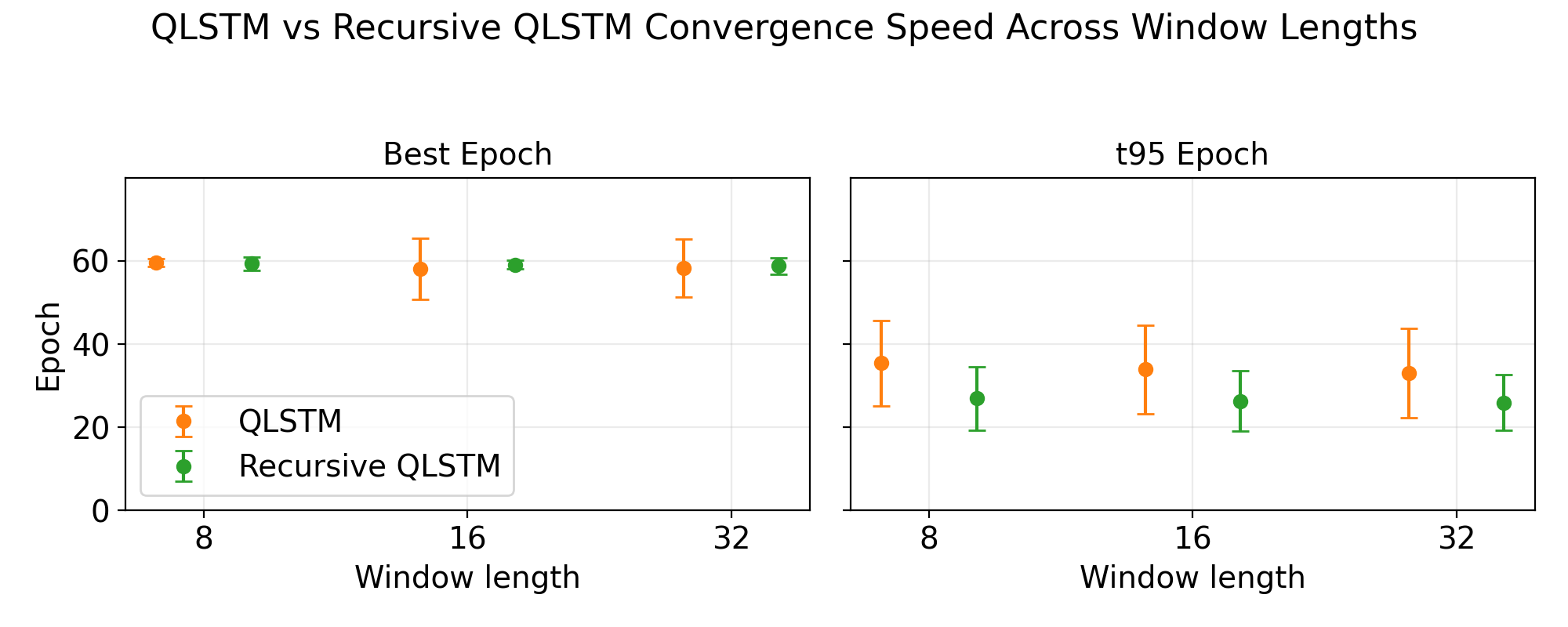}
\caption{Best-epoch and $t_{95}$ convergence statistics for QLSTM and Recursive QLSTM across window lengths of 8, 16, and 32. Error bars denote the standard deviation across 20 random seeds.}
\label{fig:convergence}
\end{figure}

Figure~\ref{fig:convergence} compares the epoch of the minimum test loss and $t_{95}$. The two models reached their minimum test loss at broadly similar epochs across the three window lengths. Recursive QLSTM reached its best test loss slightly earlier at $L=8$, slightly later at $L=16$, and at a comparable epoch at $L=32$. Thus, the best-epoch statistic does not indicate a consistent timing advantage for either architecture.

In contrast, Recursive QLSTM reaches 95\% of its eventual test-loss improvement in fewer epochs across all three window lengths. This pattern suggests that the recursive architecture reaches a useful near-optimal performance range earlier, even when the final epoch of minimum test loss occurs at a similar point in training.

At $L=16$ and $L=32$, Recursive QLSTM also shows narrower
error bars, indicating more consistent convergence across seeds.

\subsection{Predictive Accuracy}

\begin{figure}[htbp]
\centering
\includegraphics[scale=0.4]{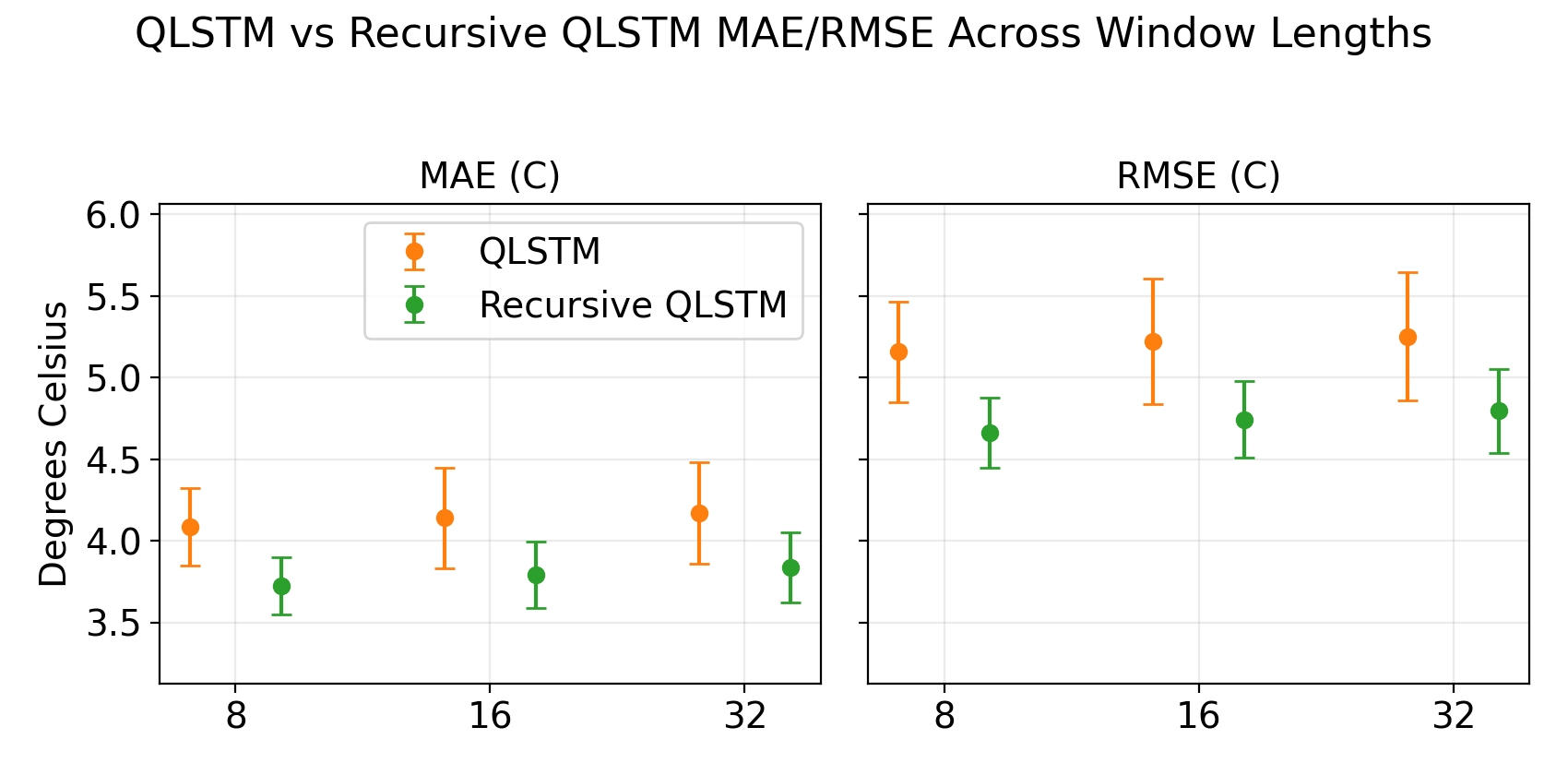}
\caption{Aggregate held-out test MAE and RMSE for QLSTM and Recursive QLSTM across window lengths of 8, 16, and 32. Metrics are evaluated at the minimum-test-loss epoch of each seed and averaged across 20 random seeds.}
\label{fig:accuracy}
\end{figure}

Figure~\ref{fig:accuracy} reports aggregate MAE and RMSE on the held-out test set, calculated at the minimum-test-loss epoch of each seed. Recursive QLSTM achieves MAE below 3.9$^\circ$C for all window lengths, compared with approximately 4.1--4.2$^\circ$C for QLSTM. The RMSE results show the same direction of effect, indicating that the recursive architecture reduces both average prediction error and the influence of larger residuals.

Recursive QLSTM remains favorable at every tested context
length, with smaller cross-seed deviations in Figure~\ref{fig:accuracy} and
Table~\ref{tab:residual_summary}.

\begin{figure}[htbp]
\centering
\includegraphics[scale=0.28]{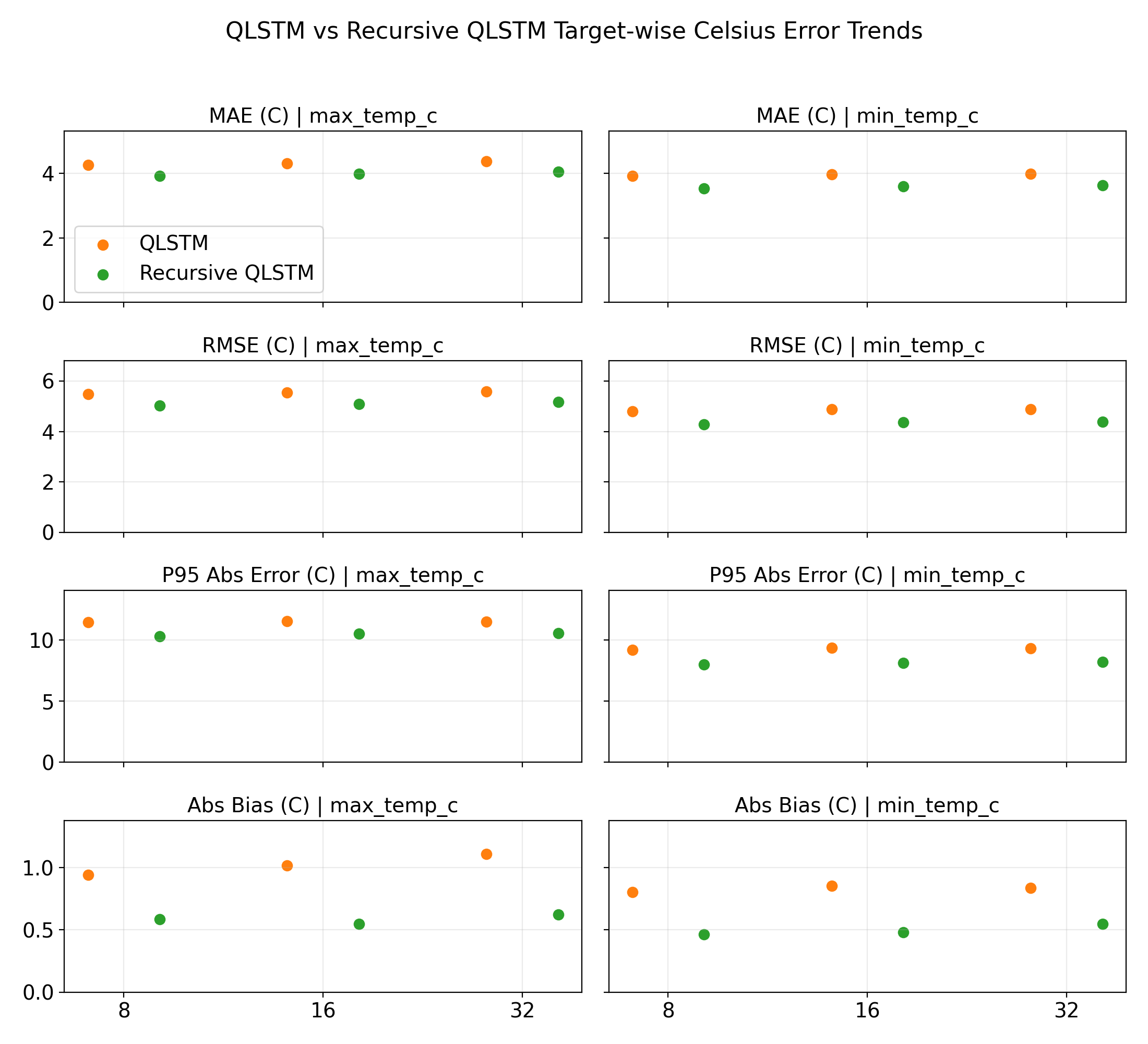}
\caption{Target-wise held-out test-error comparison for next-day maximum and minimum temperature. Metrics are evaluated at the minimum-test-loss epoch of each seed.}
\label{fig:target_accuracy}
\end{figure}

Figure~\ref{fig:target_accuracy} shows that the improvement holds for both maximum
and minimum temperature. Recursive QLSTM has lower MAE,
RMSE, P95 absolute error, and absolute bias at every window
length.

Table~\ref{tab:residual_summary} confirms lower mean and tail errors, higher $R^2$,
and smaller cross-seed variation for Recursive QLSTM.

\begin{table}[t]
\caption{Best-epoch residual robustness on the held-out test set. Values are mean $\pm$ standard deviation across 20 seeds.}
\label{tab:residual_summary}
\centering
\scriptsize
\setlength{\tabcolsep}{2pt}
\begin{tabular}{cccccc}
\toprule
\textbf{$L$} & \textbf{Model} & \textbf{MAE} & \textbf{RMSE} & \textbf{P95 AE} & \textbf{$R^2$}\\
\midrule
8 & QLSTM & $4.08\pm0.24$ & $5.15\pm0.31$ & $10.25\pm0.88$ & $0.625\pm0.046$\\
8 & Rec. QLSTM & $3.73\pm0.18$ & $4.66\pm0.21$ & $8.97\pm0.67$ & $0.69\pm0.028$\\
16 & QLSTM & $4.14\pm0.31$ & $5.22\pm0.38$ & $10.38\pm1.04$ & $0.616\pm0.059$\\
16 & Rec. QLSTM & $3.79\pm0.20$ & $4.74\pm0.24$ & $9.07\pm0.71$ & $0.685\pm0.032$\\
32 & QLSTM & $4.17\pm0.31$ & $5.25\pm0.39$ & $10.38\pm1.04$ & $0.617\pm0.060$\\
32 & Rec. QLSTM & $3.84\pm0.21$ & $4.80\pm0.26$ & $9.22\pm0.74$ & $0.681\pm0.034$\\
\bottomrule
\end{tabular}
\vspace{1mm}

\raggedright\scriptsize All error metrics are in $^\circ$C.
\end{table}

\subsection{Generalization Gap}

\begin{figure}[htbp]
\centering
\includegraphics[scale=0.35]{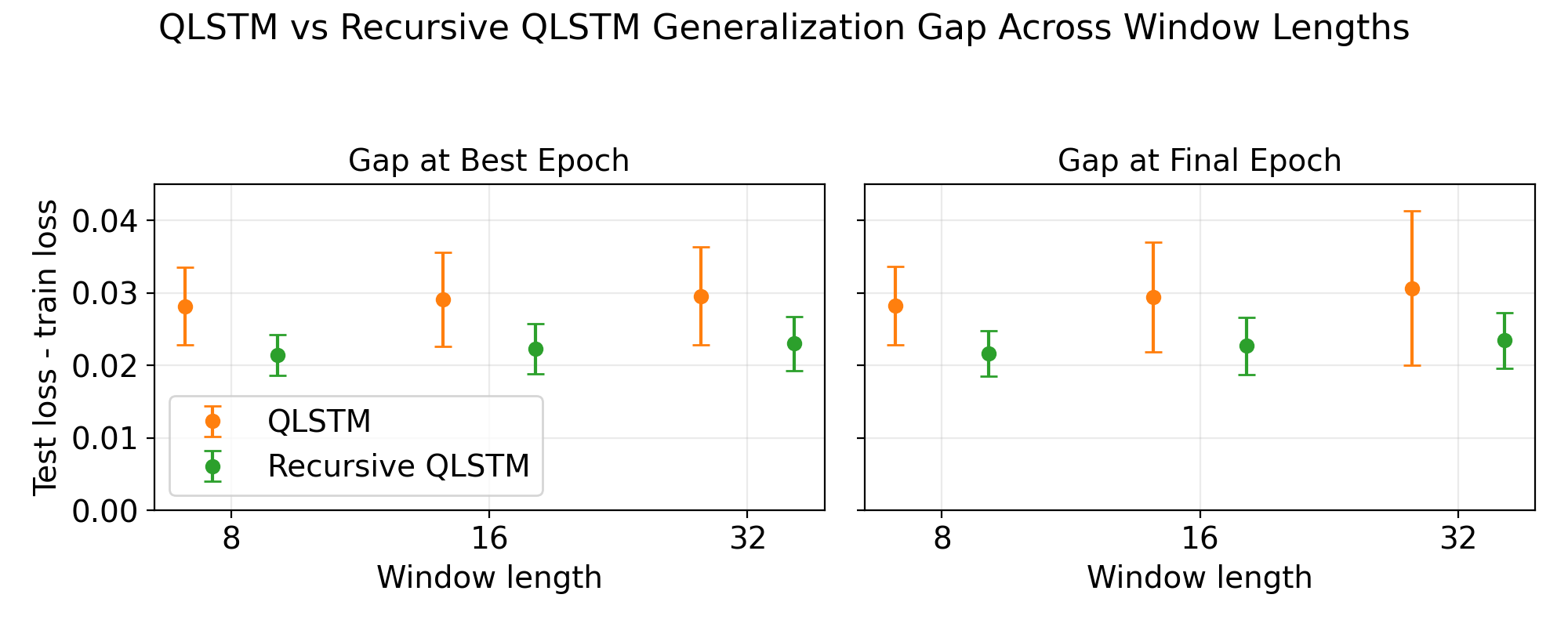}
\caption{Train--test generalization-gap comparison for QLSTM and Recursive QLSTM across window lengths of 8, 16, and 32. Results are shown at the minimum-test-loss epoch and at the final training epoch. Error bars denote the standard deviation across 20 random seeds.}
\label{fig:gap}
\end{figure}

Figure~\ref{fig:gap} shows a consistently smaller train--test loss gap for
Recursive QLSTM at both the best and final epochs, with
reductions of roughly one-quarter to one-third. Its lower
standard deviations further indicate more repeatable
out-of-sample behavior across seeds.

\section{Discussion}
The results distinguish between early practical improvement and the timing of the single lowest test-loss epoch. Recursive QLSTM reaches a near-optimal solution earlier, as reflected by a smaller $t_{95}$ across the tested window lengths. However, the epoch of the minimum test loss is broadly comparable between the two models and does not show a consistent advantage for either architecture. Reporting both measures therefore helps distinguish rapid early improvement from the point at which the best observed test loss occurs.

The contribution of this study is not only lower average forecasting error. Recursive QLSTM also shows a more consistent training pattern. Under the same set of random seeds, it reaches a useful performance range earlier, reduces both average and larger residual errors, and varies less across runs. This matters in small VQC-based experiments, where results can be affected by initialization. Better mean performance together with lower variation gives more confidence that the recursive design is contributing beyond a favorable single run.

The weather-station task also clarifies when the model is most appropriate. Daily minimum and maximum temperature are not arbitrary time-series signals: they contain strong annual seasonality, short-term persistence, and local departures caused by changing weather conditions. The input variables used here combine calendar encodings with recent temperature and degree-day information, so the model is being asked to refine a short-horizon forecast from a compact but physically meaningful context. Recursive QLSTM is most suitable for this kind of setting: one-step or short-horizon prediction, limited input dimensionality, and data where recent temporal context is informative but longer windows may introduce redundant or noisy seasonal information.

The target-wise results further suggest a useful operational characteristic. Recursive QLSTM improves both maximum and minimum temperature forecasts, including the 95th-percentile absolute error. For weather applications, this matters because a model with similar mean error but fewer large residuals is preferable for decision support: missed cold nights or unusually warm days can matter more than small average deviations. The lower generalization gap indicates that the recursive mechanism is not merely fitting the training portion more aggressively; in this dataset it transfers the learned short-term structure more consistently to held-out dates.

These findings should be interpreted cautiously. Our experiments use a single urban weather station, one forecasting horizon, shallow simulated quantum circuits, and fixed hyperparameters. Recursive QLSTM appears stable for this compact forecasting setting, but the results do not demonstrate a general quantum advantage. This is in line with recent studies showing that quantum sequential models remain sensitive to task characteristics and architectural choices \cite{chen2025telecomforecasting,chen2025batchedqlstm}. Future work should test more stations, longer horizons, noisy and missing-data settings, broader hyperparameter ranges, and stronger classical baselines. NARMA-style benchmarks may also help assess whether the observed benefits extend beyond weather forecasting \cite{suzuki2022naturalqrc}.

\section{Conclusion}

We compared QLSTM and Recursive QLSTM for one-step temperature forecasting across three window lengths and 20 random seeds. Recursive QLSTM reached a useful near-optimal test-performance range earlier, achieved lower MAE and RMSE, reduced larger residual errors, and showed less variation across runs. It also exhibited a consistently smaller train--test generalization gap, indicating more repeatable out-of-sample behavior within this controlled setting.

The epoch of the minimum test loss was broadly comparable between the two architectures, suggesting that the principal convergence advantage of Recursive QLSTM lies in faster early improvement rather than a consistently earlier final optimum. These findings suggest that recursive quantum feature transformations may improve training stability and short-horizon forecasting performance in compact hybrid recurrent models.

However, the study is limited to simulated circuits, a single weather station, one prediction horizon, and fixed hyperparameters. The results therefore do not establish a broader quantum advantage or guarantee performance gains across other datasets and tasks. Future work should evaluate additional stations, longer forecasting horizons, noisy and missing-data settings, stronger classical baselines, broader hyperparameter ranges, and noisy quantum hardware conditions.

\bibliographystyle{IEEEtran}
\bibliography{references}

\end{document}